\documentclass{article}

	\usepackage{hyperref}

\hypersetup{
    colorlinks=true,
    linkcolor=red,
    citecolor=blue,
    filecolor=magenta,      
    urlcolor=cyan,
}

    \usepackage[T1]{fontenc}
	\usepackage[utf8]{inputenc}
	\usepackage[english]{babel}
	\usepackage{amsfonts}
	\usepackage{amssymb}
	\usepackage{amsthm}
	\usepackage{enumitem}
	\usepackage{bbm}
	\usepackage{mathtools}
	\usepackage{subfigure}
	\usepackage{booktabs}	
	\usepackage{setspace}
	\usepackage{graphicx}
	\usepackage{fullpage}
	
	\usepackage{soul}
	\usepackage{xspace}
	\usepackage{csquotes}
	\usepackage{changepage}
	\usepackage{verbatim}
    \usepackage{tocbasic}
    \usepackage{placeins}
    
	\usepackage[framemethod=tikz]{mdframed}
	\usepackage{todonotes}
	\presetkeys{todonotes}{color=black!5,inline}{} 

    \usepackage{tcolorbox}
    \newtcbox{\feedback}{nobeforeafter,colframe=black,colback=white,boxrule=0.5pt,arc=2pt,
      boxsep=0pt,left=2pt,right=2pt,top=2pt,bottom=2pt,tcbox raise base}
      
    \usepackage{amsthm}
     
    \newtheorem{thm}{Theorem}

    \newtheorem{proposition}{Proposition}
    \newtheorem{lem}{Lemma}
    \newtheorem{cor}{Corollary}

    \newcommand{\parens}[1]{\left(#1\right)}
    \newcommand{\prob}[1]{\mathbb{P}\parens{#1}}

\usepackage{arxiv}
\usepackage{url}            
\usepackage{nicefrac}       
\usepackage{microtype}      
\newcommand{\indep}{\perp \!\!\!\perp}

\title{Bias In, Bias Out? Evaluating the Folk Wisdom}
\date{February 11, 2020}
\author{
    Ashesh Rambachan \\
    Department of Economics, Harvard University \\
    \texttt{asheshr@g.harvard.edu}
    \And
    Jonathan Roth \\
    Department of Economics, Harvard University \\
    \texttt{jonathanroth@g.harvard.edu}
}

\begin{document}
\maketitle

\begin{abstract}
We evaluate the folk wisdom that algorithmic decision rules trained on data produced by biased human decision-makers necessarily reflect this bias. We consider a setting where training labels are only generated if a biased decision-maker takes a particular action, and so ``biased'' training data arise due to discriminatory selection into the training data. In our baseline model, the more biased the decision-maker is against a group, the more the algorithmic decision rule favors that group. We refer to this phenomenon as \textit{bias reversal}. We then clarify the conditions that give rise to bias reversal. Whether a prediction algorithm reverses or inherits bias depends critically on how the decision-maker affects the training data as well as the label used in training. We illustrate our main theoretical results in a simulation study applied to the New York City Stop, Question and Frisk dataset.
\end{abstract}

\section{Introduction}
Algorithms have the promise to improve upon human decision-making in a variety of settings, but concerns abound that algorithms may produce decision rules that are biased against particular groups. A particular fear is that if the training data is generated by human decision-makers that discriminate against a particular group, then the algorithm will reflect this bias. This concern is captured by the common refrain ``bias in, bias out'' \cite{BarocasSelbst2016, Mayson(18)}.

In this paper, we evaluate the folk wisdom that algorithms trained on data produced by biased human decision-makers will necessarily inherit bias. Through the lens of a classic model of discrimination in economics, we consider the case where ``biased'' training data arise due to discriminatory selection into the training data and illustrate that algorithms trained over such biased training data do not necessarily inherit bias. In fact, for a common class of prediction exercises, we show that the opposite is true: The more biased the decision-maker is against a group in the training data, the more favorable the algorithm is toward that group. We refer to this phenomenon as \textit{bias reversal}. We clarify the conditions that give rise to bias reversal and discuss how alternative biases in the training data affect resulting algorithms.

We consider a baseline model with three elements that together produce bias reversal. First, we consider a setting in which labels in the training data are created only if a decision-maker chooses to take a particular action. This is commonly known as the \textit{selective labels problem} \cite{LKLLM2017, KLLLM2018}. For instance, we may only obtain data on whether a pedestrian is carrying contraband if a police officer chooses to search the pedestrian. Likewise, a college may only obtain data on a student's academic performance in college if an admissions officer chooses to accept the student, a bank may only obtain data on a borrower's creditworthiness if a loan officer chooses to grant the borrower a loan, and a firm may only obtain data on a job applicant's productivity if that applicant is hired. Second, we follow a classic literature in the economics of discrimination and assume that the decision-maker is a \textit{taste-based discriminator} against the disadvantaged group \cite{Becker1957, AltonjiBlank(99), KnowlesPersicoTodd2001, AnwarFang2006, ArnoldDobbieYang2017}. This means that the decision-maker acts as if they receive a different payoff (or face a different cost) for taking the action of interest against a particular group. This may arise due to preferences, costs, or misperceptions. As a result, bias in our model manifests itself through selection into the training data. Finally, we assume that the decision-maker has access to \textit{unobservables}, which are features that are informative about the label of interest but are unavailable in the observed training data. Each of these three elements -- selective labels, taste-based discrimination and unobservables -- are critical to bias reversal.

In this baseline model, we then show that the \textit{more biased} the decision-maker is against the disadvantaged group, the \textit{more favorable} the resulting algorithmic decision rule is toward the disadvantaged group. For example, in settings where police officers are biased in their decision to search pedestrians for contraband, an algorithmic decision rule trained to predict whether a pedestrian is carrying contraband using previously conducted searches would search fewer African American pedestrians than if police officers were unbiased in their search decisions. Similarly, in settings where managers are biased against African-Americans in hiring decisions, an algorithmic decision rule trained to predict employee performance using data on previously hired employees would hire more African American applicants than if the managers were unbiased in their hiring decisions.

To illustrate the intuition for this result, consider the example of police searches. Suppose that police assess the probability that an individual is carrying contraband, and search people with high assessed probabilities. Police base their search decision on a number of factors that are recorded in the data (the time of stop, location, demographics of the individual), as well as subjective information that is not recorded in the data (their evaluation of the individual's behavior). Because police choose to search individuals with risky behavior that is unobservable to the data scientist, an algorithm trained to predict whether contraband was found using a sample of conducted searches will tend to make predictions that are too high for the general population. However, this selection issue will be mitigated for African Americans if police officers are racially biased. Indeed, in the extreme case where police officers are so biased that they search \textit{all} African Americans, regardless of underlying risk, then there will be no selection on unobservable behavior for African Americans in the training data. Thus, the more biased are police officers, the more favorable is the training data for African Americans, and hence the more the algorithm learns to favor African Americans. 

We emphasize that our results do not imply that biased data can never produce biased algorithms. Rather, our results highlight that whether an algorithm does or does not inherit bias depends crucially on the form of the bias and the training of the algorithm. To illustrate this, we consider modifications to our baseline model that can produce effects in line with the usual ``bias in, bias out'' intuition. First, bias reversal crucially depends on the fact that the algorithm is trained to predict the outcome of interest (carrying contraband in the policing example) in the sample where the outcome is available. The typical ``bias in, bias out'' result can be obtained if either i) the algorithm is instead trained to predict the human decision, or ii) the outcome of interest is assumed to be zero for those not selected by the human decision-maker. Second, while we assume that selection into the training data is determined by a biased decision-making process, we assume that the label of interest is measured without bias. This rules out ``label bias,'' an additional source of bias in training data mentioned in the literature on algorithmic fairness -- see \cite{CorbettDaviesGoel(18)} for a discussion.\footnote{In the policing example, label bias would arise if police officers discriminated against African-Americans by fabricating evidence against them or ignoring evidence against whites.}

This paper relates to several recent works that study fairness and discrimination across computer science and the social sciences. First, several papers consider properties of algorithms that are trained on selectively-labelled data.  \cite{KLLLM2018} and \cite{LKLLM2017} define the selective labels problem and discuss its implications for evaluating the predictive performance of algorithms. \cite{KallusZhou(18)} studies how the selective labels problem impacts fairness-adjusted predictors. \cite{DeArteaga(18)} illustrates that the selective labels problem cannot be addressed via standard sample selection procedures and propose new methodology to deal with it. \cite{Cowgill2019} shows that when there are selective labels, an algorithm can improve upon human decisionmaking if the human decisions are sufficiently noisy. \cite{MadrasEtAl2019} proposes a causal modeling approach to estimating fair prediction functions in the presence of unobserved features. Finally, \cite{KannanEtAl(18)} studies the related problem of how a fairness-minded decision-maker (e.g. college admissions officer) should select a screening rule if the selected data from that screening decision are used downstream by a Bayesian decision-maker (e.g. employer). Our work is also related to a series of legal papers that have argued that automating decisions will magnify discrimination due to historical biases in existing training data -- see \cite{BarocasSelbst2016}, \cite{Chander(17)}, \cite{Mayson(18)}. In contrast, our results suggest that for certain prediction exercises, historical biases in training data can produce automated decision rules that may reverse discrimination. Conversely, our results also imply that if an algorithm is trained on data that is produced by a decision-maker that exhibits explicit affirmative action towards a group, the algorithm could, in fact, inherit bias.

Our analysis abstracts away from several potentially important considerations that could be considered in future work. First, we assume that the outcome $Y$ itself is measured without bias. This is often a significant concern in many empirical settings of interest. Second, our main theoretical results focus on properties of the optimal, population prediction function under squared loss -- i.e., the conditional expectation of the outcome given the features -- and abstracts away from finite-sample considerations. Although our simulation evidence indicates that our results still hold in finite-sample, this deserves further attention. Similarly, extending these results to more general loss functions may be of interest. Third, our results focus on an algorithmic decision rule that is trained ``naively'' by the data-scientist, meaning that they do not adjust for selection into the data nor impose any additional fairness criteria. Finally, we focus attention on a taste-based model for discrimination. Other models of discriminating behavior may yield different conclusions. For example, discrimination may arise due to stereotypes (e.g. \cite{ShleiferEtAl(16)}) or differential noise in the decision-maker's predictions across groups (e.g. \cite{Li(17)}).

The remainder of this paper is structured as follows. Section \ref{section:model} presents our baseline model. Section \ref{section:main-results} states our baseline results and Section \ref{section: extensions} discusses extensions. Section \ref{section:application} illustrates our results in simulations based on New York Stop, Question and Frisk data. We place more involved proofs in the Appendix. 

\section{A Model of Biased Decisions}\label{section:model}
In this section, we develop a model wherein the training data given to a predictive algorithm is generated by a biased human decision-making process. For the sake of exposition, we discuss the model in the context of police bias in pedestrian searches and refer to the decision-maker as the police throughout. This will more clearly connect our theoretical results with our empirical application to New York Stop, Question and Frisk. However, this model is broadly applicable to other settings with selective labels such as college admissions, loan decisions, and hiring decisions, among many others. We discuss the connection to these other settings in Section \ref{section: fewer labels}.

Police officers wish to search individuals that have a high probability of carrying contraband. Following \cite{Becker1957} and a large literature in economics, police officers are taste-based discriminators against African Americans.\footnote{Unlike \cite{AnwarFang2006} and \cite{KnowlesPersicoTodd2001}, we do not assume that the individual's decision to carry contraband responds to police search decisions. As a result, we do not introduce an equilibrium concept such as Nash equilibrium.} Based on the search decisions of police officers, data are then revealed to the data scientist. If a police officer searches an individual, the data scientist observes the result of that search (was the individual carrying contraband?), some characteristics of the individual and the stop (age, gender, location of stop, time of stop, etc.) as well the race of the individual. The data scientist then uses this training data to construct an algorithm to predict which individuals are most likely to be carrying contraband. We focus our attention on analyzing properties of the predictive algorithm produced by the data scientist.

\subsection{The population}
Individuals in the population are characterized by the random vector $(X, U, R, Y)$. Let $X \in \mathcal{X}$ denote some set of characteristics about the individual that are typically recorded after a police search such as age, gender, location of stop, time of stop, etc. Let $U \in \mathcal{U}$ denote characteristics of an individual that are observed by a police officer prior to a search but are typically not recorded. For example, this may consist of the police officer's evaluation of the individual's behavior prior to the stop or the individual's behavior during the stop. Importantly, $U$ is observed by the police officer but is unobserved to the data scientist. Finally, $R \in \{0, 1\}$ denotes the race of the individual with $R = 1$ for African Americans, and $Y \in \{0, 1\}$ denotes whether the individual is carrying contraband. The population is then described by the joint distribution $\mathbb{P}$ of the random vector $(Y, X, R, U)$.

\subsection{Police decisions}
The police observe the characteristics $(X, U, R)$ of each individual and decide whether to search that individual. The police officer receives a positive payoff $b > 0$ if they find contraband after searching an individual and without loss of generality, we normalize this payoff to one, $b = 1$. The police officer receives a payoff of zero if the individual is not searched. The police officer incurs a cost $c > 0$ for every search. 

In addition, the police are taste-based discriminators against African Americans and receive an additional payoff $\tau > 0$ from searching African Americans. The parameter $\tau$ parametrizes the degree to which the police are biased against African Americans. The larger the magnitude of $\tau$, the more biased the police are against African Americans. In sum, the police's payoffs from conducting a search are $Y + \tau R - c $. In order to maximize their expected payoff, the police will decide whether to search according to a threshold rule:
    $$S^*(X, U, R) = 1\left( \mathbb{E}[Y | X, U, R] \geq c - \tau \cdot R \right). $$
Notice that the bias of the police implies that a lower threshold for search is applied to African Americans. In this sense, the police are biased against African Americans. For simplicity, we assume for now that the police make their decisions based on an optimal prediction of $Y$ given $(X,U,R)$ under squared loss (i.e., $\mathbb{E}[Y|X,U,R]$), although we show in Section \ref{subsec:noisy decisionmaking} that our main results extend to the case where the police use noisy but unbiased predictions. 

\subsection{The prediction problem}
The data scientist then observes data consisting of individuals that are stopped by the police. There are ``selective labels'' -- the data scientist only observes whether an individual was carrying contraband ($Y$) if the police searched the individual ($S^* = 1$). The data-scientist thus observes the pair $(Y,X,R, S^*)$ for those with $S^* = 1$. In some of our results, we will also consider what happens if the data scientist is able to observe $(X, R, S^*)$ but not $Y$ for those who are not searched by the police. Let $\hat{\mathbb{P}}_{\tau}$ denote the joint distribution of the data that is revealed to the data scientist. We index the probability distribution of the observed data by the police's discrimination parameter $\tau$ as our results will focus on comparative statics over $\tau$.

Using the observed data, the data scientist constructs a predictive algorithm of whether an individual is carrying contraband $Y$ using the observed features $(X, R)$. In our baseline model, we suppose that the data scientist trains the algorithm using only the data where the outcome is available ($S^* = 1$). For simplicity, we abstract from the estimation problem and consider properties of the optimal predictor under squared loss, $\mathbb{E}_{\hat{P}_{\tau}}[Y | X, R, S^* = 1]$, where $\mathbb{E}_{\hat{P}_{\tau}}[\cdot]$ denotes the conditional expectation over the distribution of observed data. Note that for now we suppose that race is included as a feature; we discuss relaxing this assumption in Section \ref{section: group membership}.

\section{Baseline results}\label{section:main-results}
\subsection{Bias reversal}

We now present our bias reversal result, which examines how an algorithm trained to predict $Y$ in the searched sample ($S^* = 1$) can reverse bias. We first sketch the intuition and then formally state the result.

Since the police incorporate the unobservable $U$ into their search decision, the training data of conducted searches will tend to be composed of individuals that have values of $U$ associated with higher probability of $Y=1$. As a result, the predictive algorithm trained on the selected training data will tend to over-predict the label $Y$ for the whole population. However, as the police officers become more biased, this selection problem becomes less severe for African Americans. Intuitively, the more biased are the police officers against African Americans, the more likely they are to search any given African American, and so there is less selection on the unobservable $U$. In the extreme case where $\tau \geq c$, police officers search all African Americans, and there is no selection on the unobservable $U$ for African Americans. The predictive algorithm thus becomes more favorable to African Americans as the police officers become more biased. 

\begin{thm}\label{theorem:comparative-static}
    $\mathbb{E}_{\hat{P}_\tau}[Y | X =x , R = 1, S^* =1]$ is weakly decreasing in $\tau$ for all $x \in \mathcal{X}$ and $\tau$ such that $\hat{P}_{\tau}(S^* = 1 \,|\, X = x, R = 1) > 0$. Likewise, $\mathbb{E}_{\hat{P}_\tau}[Y | X =x , R = 0, S^* = 1]$ is constant in $\tau$ for all $x \in \mathcal{X}$ and $\tau$ such that $\hat{P}_{\tau}(S^* = 1 \,|\, X = x, R = 0) > 0$.
\end{thm}
\noindent \textit{Proof.} Define $\mu_{X,R,U} := \mathbb{E}[Y | X, R, U]$ and $\mu_{X,R} := \mathbb{E}[Y | X, R]$. Let $U^* = \mu_{X,R,U} - \mu_{X,R}$, so that $\mu_{X,R,U} = \mu_{X,R} + U^*.$ Note that $S^* = 1$ if and only if $U^* \geq T(X, R, \tau)$ for the threshold $T(X, R, \tau) = (c - \tau \cdot R) - \mu_{X,R}$. Applying the law of iterated expectations,
    \begin{align*}
        \mathbb{E}\left[Y | X=x, R =r, S^* = 1\right] = \mathbb{E}\left[Y | X=x, R=r, U^* \geq T(x,r, \tau) \right] \\
        = \mathbb{E}\left[ \mathbb{E}\left[Y | X=x, R=r, U\right] |  X=x, R=r, U^* \geq T(x,r,\tau)\right] \\
        = \mu_{x,r} + \mathbb{E}\left[ U^* |  X=x, R=r, U^* \geq T(x,r,\tau) \right].
    \end{align*}
\noindent Note that for $r=1$, $T(x,r,\tau)$ is decreasing in $\tau$. It follows immediately that $E[ U^* |  X=x, R=r, U^* \geq T(x,r,\tau) ]$ is weakly decreasing in $\tau$, which gives the first desired result. Likewise, when $r=0$, $T(x,r,\tau)$ does not depend on $\tau$, which gives the second result. $\Box$

Theorem \ref{theorem:comparative-static} shows that as the police become more biased against African Americans, the predictions of the algorithm trained on the selected data become more favorable to African Americans, in the sense that African Americans are predicted to have lower risk of carrying contraband. This implies the more biased the police are against African Americans, the fewer African Americans will be searched by an automated search rule that uses these predictions.

\begin{cor}\label{corollary:automated-search}
    Consider the automated search rule:
        $S^{automated}_{\tau}(x, r) = 1\left(\mathbb{E}_{\hat{\mathbb{P}}_\tau}[Y | X = x, R = r, S^* =1] \geq c_{min} \right)$ for some $c_{min}\in [0,1]$.\footnote{We implicitly assume that $P(S^*=1 \,|\, X =x , R=r) > 0$ for almost every $(x,r)$, so that the search rule is well-defined.} Then $S^{automated}_\tau(x,1) \leq S^{automated}_{\tau'}(x,1) $ for any $\tau' < \tau$, so any African-American searched under $\tau$ is also searched under $\tau'$, whereas $S^{automated}_{\tau}(x,0)$ does not depend on $\tau$.
    It follows that the fraction of African Americans searched under $S^{automated}_\tau$ (i.e, $\mathbb{E}[ S^{automated}_\tau(X,R) \,|\, R = 1]$) is decreasing in $\tau$, whereas the fraction of whites searched under $S^{automated}_\tau$ is constant in $\tau$.    
\end{cor}

Corollary \ref{corollary:automated-search} states that an automated decision rule based on a threshold rule using $\mathbb{E}_{\hat{\mathbb{P}}_\tau}[Y | X = x, R = r, S^* =1]$ searches fewer African-Americans the larger is the bias $\tau$ in the training data.

These results clarify the manner in which the bias of police officers influences the algorithmic treatment of African Americans under an automated search rule. We do not take a stance directly on whether the algorithm's treatment of African Americans for any given $\tau$ is ``fair'' in a formal sense.\footnote{Results in \cite{KLR(16)} highlight that an algorithm cannot simultaneously satisfy several common definitions of fairness if the base rates of risk differ across groups.} However, any sensible notion of fairness would suggest that if a given decision rule is unfair to African Americans, then any decision rule that is ``harsher'' to African Americans (i.e. more likely to search any given African American) and treats whites the same is at least as unfair. Therefore, Theorem \ref{theorem:comparative-static} and Corollary \ref{corollary:automated-search} suggest that if a decision rule which is based upon a prediction function trained on data produced by police officers that discriminate against African Americans ($\tau > 0$) is unfair, then a decision rule which is based upon a prediction function trained on data produced by police officers that are unbiased against African Americans ($\tau = 0$) would be even \textit{more} unfair to African Americans.


It is important to note that while we presented these results in the context of police searches, they apply to a broader class of settings in which there is a selective labels problem, the decision-maker that produces the selective labels is a taste-based discriminator against a particular group, and the discriminator has access to unobservables. For example, in Section \ref{section: fewer labels}, we discuss how these results extend to other settings such as loan applications, hiring decisions and college admissions. When these conditions hold, the more biased is the decision-maker  against a group, the more the algorithmic decision-rule favors that group. We refer to this phenomenon as ``\textit{bias reversal},'' which captures the sense in which the algorithm's predictions reverses the taste-based discrimination of the decision-maker. 


\subsection{Bias inheritance for alternative prediction exercises}
In Theorem \ref{theorem:comparative-static} and Corollary \ref{corollary:automated-search}, we assumed that the data scientist constructs an algorithm to predict the observed label $Y$ using the training data for the searched sample ($S^* = 1$). We now consider what happens if a different label and sample is used. First, the data scientist may instead predict the human decision $S^*$ itself over the full population. This is a common type of prediction problem in some contexts. For example, a series of papers note that using the human decision as the label is common in training algorithms to automate hiring decisions \cite{Cowgill2018, Cowgill2019, RaghavanEtAl(19)}. For this prediction exercise, bias reversal no longer holds. Instead, the comparative static now implies that the prediction function inherits bias --- as the police become more biased against African-Americans, the predictions of the algorithm trained in this way become less favorable to African-Americans.

\begin{thm}
    \label{thm: comparative static for expectation of s-star}
    $\mathbb{E}\left[S^* | X =x, R=1\right]$ is weakly increasing in $\tau$ for all $x \in \mathcal{X}$. 
\end{thm}
\noindent \textit{Proof.} By the law of iterated expectations, $\mathbb{E}[ S^* | X = x, R =1] = \mathbb{E}[ \mathbb{E}[ S^* \,|\, X = x, U, R =1] ]$. Then,
    \begin{align*}
        \mathbb{E}[ S^* | X=x, R=1] &= \int_{u \in \mathcal{U}} \mathbb{E}[ S^* | X=x, U = u, R=1 ] \, dF(u)   \\
        &= \int_{ \{u \in \mathcal{U} : S^*(x,u,1) = 1 \} } dF(u) \\
        &= \int_{ \{u \in \mathcal{U} : \mathbb{E}[Y|X=x,U=u,R= 1] \geq c - \tau\} }  dF(u).
    \end{align*}
\noindent It follows that for $\tau_1 < \tau_2$,
    \begin{equation*}
        \mathbb{E}[ S^* | X=x, R=1, \tau = \tau_2] - \mathbb{E}[ S^* | X=x, R=1, \tau = \tau_1] = \int_{u \in \mathcal{U}_{12}} dF(u),
    \end{equation*}
\noindent for $\mathcal{U}_{12} = \{u \in \mathcal{U} : c - \tau_2 \leq \mathbb{E}[Y|X=x,U=u,R= 1] \leq c - \tau_1 \}$, which gives the desired result. $\Box$

A second alternative prediction exercise that the data scientist may also consider is to predict the compound outcome that the individual was searched by the police and that the individual was carrying contraband. That is, construct an algorithm to predict the label $Y \cdot S^*$ over the full sample. Put otherwise, the data scientist imputes the missing label $Y$ to be zero if $S^* = 0$. This type of prediction exercise is common in certain medical applications (see, e.g. \cite{mullainathan2017does}). In this case, we again find that the prediction function inherits bias from the police officers' search decisions. 

\begin{thm}
    \label{thm: comparative static for y times s}
    $\mathbb{E}\left[ Y S^* | X = x, R =1 \right]$ is weakly increasing in $\tau$ for all $x \in \mathcal{X}$.
\end{thm}

\noindent \textit{Proof.} The proof of this result is analogous to Theorem \ref{thm: comparative static for expectation of s-star}. $\Box$

The key distinction between these alternative prediction exercises and our earlier result is that bias now drives a wedge between the true outcome of interest and the label that the algorithm is trained on ($S^*$ or $Y \cdot S^*$), but the human bias does not affect sample composition. By contrast, in the original setting that predicts $Y$ over the selected sample with $S^* = 1$, the bias affects the prediction exercise only through sample composition. This is a crucial yet subtle difference. 

Theorem \ref{thm: comparative static for expectation of s-star} and Theorem \ref{thm: comparative static for y times s} immediately imply that an automated decision rule that is based upon predictions of $S^*$ or $YS^*$ will inherit bias --- that is, search more African Americans as police officers become more biased. 

\begin{cor}
 Consider the automated search rule $\check{S}^{automated}_{\tau}(x, r) 1\left( \hat{Y}(x, r) \geq c_{min} \right)$ for some $c_{min} \in [0, 1]$, where $\hat{Y}(x, r) = \mathbb{E}\left[ S^* \,|\, X = x, R = r \right]$ or $\hat{Y}(x, r) = \mathbb{E}\left[ Y S^* \,|\, X = x, R = r \right]$. Then, $\check{S}_{\tau}^{automated}(x, 1) \leq \check{S}_{\tau'}^{automated}(x, 1)$ for $\tau < \tau'$, so any African American that is searched under $\tau$ is also searched under $\tau'$. It follows that the fraction of African Americans searched under $\check{S}_{\tau}^{automated}$ (i.e. $\mathbb{E}\left[\check{S}^{automated}(X, R) \,|\, R = 1\right]$) is increasing in $\tau$. 
\end{cor}


Taken together, these results show that the choices of label ($Y$ vs. $S^*$ vs. $Y \cdot S^*$) and training sample ($S^* =1$ vs. full sample) play a key role in determining whether human biases propagate into algorithmic predictions and automated decisions, formalizing an argument made heuristically in \cite{KLMS(18)}. Table \ref{tbl:prediction_setups} summarizes our results across the three prediction exercises considered.

\begin{table}[h]
    \centering
    \caption{Summary of prediction exercises}
    \begin{tabular}{l l l}
        Outcome & Training sample & Comparative static \\
        \hline
         $Y$ & $S^* = 1$ & Bias reversal \\
         $S^*$ & Full sample & Bias inheritance \\
         $Y \cdot S^*$ & Full sample & Bias inheritance
    \end{tabular}
    \label{tbl:prediction_setups}
\end{table}

\section{Extensions}\label{section: extensions}

\subsection{When discrimination yields fewer labels for the disadvantaged group}\label{section: fewer labels}

In other settings of interest with selective labels such as loan applications, hiring decisions, and college admissions, \textit{fewer} labels are generated when the decision-maker is biased. For example, if a hiring manager is biased against African Americans, fewer African American applicants are hired. A simple extension of our baseline model shows that an analogous comparative static still holds: the more biased the decision-maker is against a group, the more the resulting algorithmic decision rule favors that group.

As an example, consider a hiring manager that predicts the productivity $Y$ of job applicants using features $X$ that are observable to the data scientist and features $U$ that are unobservable to the data scientist. A biased hiring manager applies a higher predicted-productivity threshold for African Americans than for whites. This means the more biased is the hiring manager, the fewer African-Americans will enter the training data. However, the African-Americans who do enter the training data will be more positively selected on $U$ (i.e., on unobservables positively correlated with productivity). Thus, the more biased is the hiring manager against African-Americans, the higher will be the algorithm's predicted productivity for African-Americans and the more African Americans will be hired by an algorithmic hiring rule.

Formally, consider a modified selection rule $\tilde{S}(X, U, R) = 1\left(\mathbb{E}[Y | X, U, R] \geq c + \tilde{\tau} R\right)$, with $\tilde{\tau} > 0$. Define $\hat{P}_{\tilde \tau}$ to be the joint distribution of the data revealed to the data scientist under the selection rule $\tilde{S}(X, U, R)$. Note that this model is equivalent to the model considered earlier with $\tau = -\tilde{\tau}$. We thus obtain the immediate corollary to Theorem \ref{theorem:comparative-static}.
    \begin{cor}
        $E_{\hat{P}_{\tilde\tau}}[Y | X =x , R = 1, \tilde{S} = 1]$ is weakly increasing in $\tilde \tau$ for all $x$,$\tilde\tau$ such that $\hat{P}_{\tilde\tau}(\tilde{S} = 1 | X=x, R=1 ) >0$, while $E_{\hat{P}_{\tilde\tau}}[Y | X =x , R = 0, \tilde S = 0]$ is constant in $\tilde \tau$, for all $x,\tilde\tau$ such that $\hat{P}_{\tilde\tau}(\tilde{S} = 1 | X=x, R=0 ) >0$.
        
        Moreover, consider the automated hiring rule:\footnote{We implicitly assume that $P(\tilde S=1 \,|\, X =x , R=r) > 0$ for almost every $(x,r)$, so that the hiring rule is well-defined.}
        $$\tilde{S}^{automated}_{\tilde{\tau}}(x, r) = 1\left(\mathbb{E}_{\hat{\mathbb{P}}_{\tilde\tau}}[Y | X = x, R = r, \tilde{S} =1] \geq c_{min} \right)$$
        for $c_{min} \in [0, 1]$. Then, $\tilde{S}^{automated}_{\tilde{\tau}}(x, 1) \leq \tilde{S}^{automated}_{\tilde{\tau}'}(x, 1)$ for any $\tilde{\tau} < \tilde{\tau}'$. It follows that the fraction of African Americans hired under $\tilde{S}^{automated}$ (i.e, $E[ \tilde{S}^{automated}(X,R) \,|\, R = 1]$) is increasing in $\tilde\tau$.
    \end{cor}

    
\noindent In unpacking this result, it is useful to distinguish between statistical bias and ``favorability'' of the algorithm. As the decision-maker becomes more biased, the predictions of the algorithm become more biased in a statistical sense, meaning that the magnitude of $E[ Y \,|\,X, R=1, \tilde{S}=1 ] - E[ Y \,|\,X, R=1 ]$ becomes larger. However, this statistical bias works in a way that makes algorithmic decision rules more likely to select members of the discriminated against group.

These results highlight that the phenomenon of bias reversal is not dependent on the selective labels problem leading to more labels to be collected for the disadvantaged group, and is thus applicable to a range of settings with selective labels.

\subsection{Noisy decision-making\label{subsec:noisy decisionmaking}}
We next show that the results in Section \ref{section:main-results} are robust to allowing for random noise in the officer's decisions. In the baseline model, we assumed that the police officers are able to correctly combine the available information to construct accurate predictions about risk, $\mathbb{E}[Y | X, R, U]$ and thereby rank order individuals correctly. Extensive work in the social sciences suggest that this does not hold in many applications of interest. For example, \cite{KLLLM2018} suggest that even experienced judges are unable to accurately predict recidivism in bail decisions. In the following result, we show that the comparative static in Theorem \ref{theorem:comparative-static} still holds if police officers have independent random noise in their risk assessments.

\begin{proposition}\label{prop:noise}
Suppose police search according to
    $$S^{noise}(X, U, R, \epsilon) = 1\Big( \mathbb{E}[Y | X, U, R] + \epsilon \geq c - \tau \cdot R\Big),$$
for a random prediction error $\epsilon$, where the distribution $\epsilon \,|\, X, R$ has strictly increasing hazard and $\epsilon \indep (Y,U)  \,|\, (X, R)$. Then, $\mathbb{E}_{\hat{P}_{\tau}}[Y | X = x, R = 1, S^{noise} =1]$ is weakly decreasing in $\tau$ for all $x \in \mathcal{X}$ and $\tau$ such that $\hat{P}_{\tau}(S^{noise} = 1 \,|\, X = x, R = r) > 0$.
\end{proposition}

\noindent \textit{Proof.} See Appendix for proof. $\Box$

\noindent Similarly, the comparative statics derived for the alternative prediction exercises are also robust to noisy decision-making. 

\begin{proposition}\label{prop:noise-alternative}
    The conclusions of Theorem \ref{thm: comparative static for expectation of s-star} and Theorem \ref{thm: comparative static for y times s} hold replacing $S^*$ with $S^{noise}$.
\end{proposition}

\noindent \textit{Proof}. The proof is analogous to the proofs of Theorem \ref{thm: comparative static for expectation of s-star} and Theorem \ref{thm: comparative static for y times s}, replacing expectations over $U$ with expectations over the joint distribution of $(U, \epsilon)$. $\Box$

\subsection{Excluding group membership from the predictive algorithm}\label{section: group membership}
We now consider what happens if the data scientist is forbidden from using group status in the predictive algorithm. For example, it may be illegal for a predictive algorithm to explicitly use race as a feature \cite{KLMS(18), GillisSpiess(19)}. In this case, the prediction function in the baseline model now takes the form $\mathbb{E}_{\hat{\mathbb{P}}_{\tau}}[Y | X, S^* =1]$.

Whether the comparative static in bias still holds now depends on whether group status $R$ is ``reconstructable'' from the observed features $X$. That is, it depends on whether group status is predictable from the observed features. If group status is perfectly reconstructable, then these results trivially hold for a prediction function that does not use group status as $\mathbb{E}_{\hat{\mathbb{P}}_{\tau}}[Y | X, S^* = 1] = \mathbb{E}_{\hat{\mathbb{P}}_{\tau}}[Y | X, R, S^* =1]$.

If group status is not perfectly reconstructable, then one can construct examples in which the gap in average predictions across groups for a group-blind algorithm moves in the opposite direction as the gap in average predictions across groups for an algorithm that includes race. The direction of the effect will depend on whether the marginally searched individual in the $R = 1$ group is more ``similar'' to the average person with $R = 0$ or $R=1$. As a simple example to illustrate this, suppose there is only one observed, binary feature $X$. Suppose that among whites, $X = 1$ with probability $1 - \epsilon$ for some small $\epsilon \geq 0$. Among African Americans, $X = 0$ with probability $1 - \epsilon$. Then, if the marginally searched African American has feature $X = 1$, then an increase in the bias of police officers will have a larger effect on the average prediction for whites than African Americans, as there are relatively more whites among the group with $X = 1$ in the observed data. Conversely, if the marginally searched African American has feature $X = 0$, then it will have a larger effect on the average prediction for African Americans than whites. The same intuition holds for the alternative prediction exercises that we considered earlier.

The reconstruction problem has been discussed at length elsewhere -- see, among many others, \cite{KLMR(18), LiptonEtAl(18), ChenEtAl2019, DattaEtAl(17)}. Typically, it is thought that if race is reconstructable from other features, then algorithms will exhibit bias or discriminate against minority groups. Our results illustrate that this is not true generally. If group status is reconstructable, then an algorithm that is blind to group status may exhibit bias reversal (Theorem \ref{theorem:comparative-static}). 

\section{Application: New York City Stop, Question and Frisk}\label{section:application}
We now apply these results to the New York Stop, Question and Frisk (SQF) data. We synthetically create a training data set that is produced by biased search decisions and illustrate the key comparative statics described in Section \ref{section:main-results}.

\subsection{Data description}
SQF was a program in New York City that allowed the police to temporarily stop, question, and search individuals on the street. We use publicly available, stop-level data that contains information on all stops conducted as part of the SQF program from 2008-2013, totalling over 4 million stops of pedestrians and over 350,000 searches \cite{GoelEtAl2016}.

For each recorded stop, we observe whether the stopped individual was searched for contraband and if so, an indicator for whether contraband was found. The data also contains several detailed characteristics of the stopped individual and the circumstances of the stop. The features in the data include the stopped individual's age, gender, and build, and the time and location of the stop. We treat these as the observable features $X$. Importantly, we also observe the race $R$ of the stopped individual. For simplicity, we restrict attention to stops of non-Hispanic whites and African Americans. 

The data also include the officer's stated reason for conducting the stop -- for example, the officer can select that the stop was conducted because the pedestrian was "carrying a suspicious object" or "displayed behavior indicative of a drug transaction." We treat these responses as the unobservable features $U$ that are available to the officer at the time of the search decision but are unavailable to the data scientist. This is analogous to ``soft information'' about the individual that may be available to the officer at the time of the stop but may be unavailable in certain data sets.

\subsection{Simulation design}
We conduct a simulation exercise that trains an algorithm to predict whether a stopped individual is carrying contraband on synthetic training datasets that are generated from the original SQF data. Across synthetic training datasets, we vary the degree of bias against African Americans in search decisions by selectively ``undoing'' observed searches. We then examine how changing the degree of bias against African Americans affects the resulting algorithm's predictions.

More concretely, we first subset the data to only include stops in which searches were conducted ($S^*=1$). We then randomly split the searched SQF stops into two partitions. In the first partition, we construct a predictor for carrying contraband among stops with searches. The predictor estimates $\mathbb{E}[Y | X, R, U, S^*=1]$, where $X$ is a feature vector that includes demographic information about the stopped individual such as age, gender and build as well as the location and time of the stop, and $U$ is the officer's stated reason for the stop. We construct the predictor using logistic regression, matching the approach of previous research using this data \cite{GoelEtAl2016, KallusZhou(18)}. 

In the held-out partition, we then use the estimated prediction function to construct a synthetic search flag $\hat{S}$. For individuals with $\hat{Y} = \hat{\mathbb{E}}[Y | X, R, U, S^*=1] \leq c_{R}$, we set $\hat{S} = 0$ and treat them as if they had not been searched. For individuals with $\hat{Y} > c_{R}$ for $R \in \{0, 1\}$, we set $\hat{S} = 1$. This produces a synthetic dataset at the search thresholds $(c_0, c_1)$ in which we observe $(Y, X, R, \hat{S})$ for each observation. Finally, we re-estimate the prediction function over the synthetically searched observations. We estimate the functions $\mathbb{E}[Y | X, R, \hat{S} = 1]$, $\mathbb{E}[\hat{S} | X, Y]$ and $\mathbb{E}[Y \hat S | X, R]$ using logistic regression and examine properties of the estimated prediction functions. 

We repeat this simulation for a variety of different thresholds $c_0, c_1$ to construct a series of synthetically searched observations at different levels of bias, defined as $\tau = c_0 - c_1$, against African Americans. We vary $c_0, c_1$ so that 50 percent of the synthetic dataset is always searched and only the composition of searches between African Americans and whites vary. We vary the fraction of searches that are conducted on African Americans from 80 percent to 95 percent. 

\subsection{Simulation results}
\begin{figure}[h]
    \centering
    \caption{NYC SQF Simulation Results}
    \includegraphics[width=.5\textwidth]{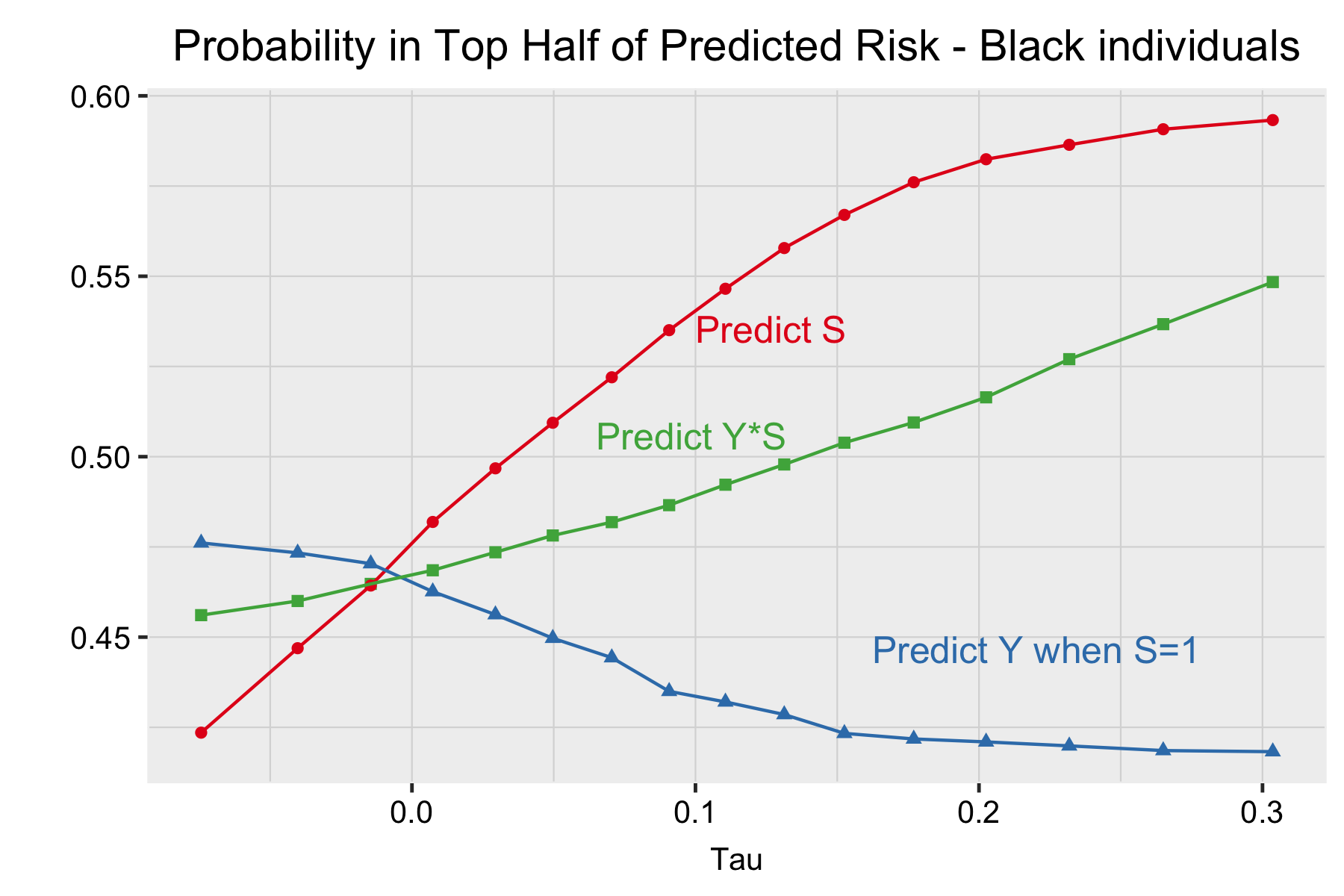}
    \label{fig:NYC-SQF-sim}
\end{figure}

Figure \ref{fig:NYC-SQF-sim} plots the results from our simulation exercise. The X-axis plots the discrimination parameter $\tau = c_0 - c_1$ across synthetic datasets. Larger values of $\tau$ correspond with a search rule that is more biased against African Americans. The Y-axis plots the fraction of African Americans that fall in the top 50 percent of predicted risk using the prediction function estimated over the synthetic dataset.

The predictions from our earlier results in Section \ref{section:main-results} hold sharply. First, as the police become more biased against African Americans, the prediction function $\hat{\mathbb{E}}[Y | X, R, \hat{S} = 1]$ becomes more favorable to them. In particular, fewer African Americans fall in the top half of predicted risk as $\tau$ increases. This illustrates our result of bias reversal in a concrete application of interest. Second, as the police become more biased against African Americans, the prediction functions $\hat{\mathbb{E}}[\hat{S} | X, R]$ and $\hat{\mathbb{E}}[Y \hat{S} | X, R]$ become less favorable to African Americans. As $\tau$ increases, more African Americans fall in the top half of predicted risk. Once again, for these prediction functions, ``bias in'' implies ``bias out.'' 

\section{Conclusion}\label{section:conclusion}
In this paper, we evaluated the folk wisdom that algorithmic decision rules trained on data that are produced by biased human decision-makers will necessarily inherit this bias. We showed that in an important class of prediction exercises, the opposite holds: The more biased the decision-maker towards a group, the more favorable is the algorithm towards that group. We called this phenomenon ``\textit{bias reversal}.'' We then showed that an important determinant of whether one obtains bias reversal or ``bias in, bias out'' is whether the human bias affects sample selection or the measured label. 

These results suggest that when we consider whether algorithms will inherit human biases, it is important to think carefully about the form of the human bias, how it affects the training sample, as well as how the labels and features are selected for the algorithm. Additionally, while some of the literature on fairness in algorithms has focused on blinding algorithms from group membership, our results suggest that there are cases in which allowing the algorithm to use race may reduce racial discrimination by the algorithm. This is in line with results found by \cite{DworkEtAl(12), KLMR(18), GillisSpiess(19)}.

\subsubsection*{Acknowledgements}
We are grateful to Isaiah Andrews, Talia Gillis, Ed Glaeser, Nir Hak, Nathan Hendren, Larry Katz, Jens Ludwig, Sendhil Mullainathan, Paul Novosad, Aaron Roth, Ben Roth, Hannah Shaffer, Jann Spiess, as well as seminar participants at Harvard, the Federal Reserve Bank of Philadelphia, and Quantco for valuable comments and feedback. We gratefully acknowledge financial support from the NSF Graduate Research Fellowship under Grant DGE1745303 (Rambachan) and Grant DGE1144152 (Roth).

\singlespacing
\bibliographystyle{unsrt}  
\bibliography{Bibliography.bib} 

\begin{thebibliography}{10}

\bibitem{BarocasSelbst2016}
Solon Barocas and Andrew Selbst.
\newblock Big data's disparate impact.
\newblock {\em California Law Review}, 104, 2016.

\bibitem{Mayson(18)}
Sandra Mayson.
\newblock Bias in, bias out.
\newblock {\em The Yale Law Journal}, 128(8):2122--2473, 2018.

\bibitem{LKLLM2017}
Himabindu Lakkaraju, Jon Kleinberg, Jure Leskovec, Jens Ludwig, and Sendhil
  Mullainathan.
\newblock The selective labels problem: Evaluating algorithmic predictions in
  the presence of unobservables.
\newblock {\em KDD '17 Proceedings of the 23rd ACM SIGKDD International
  Conference on Knowledge Discovery and Data Mining}, pages 275--284, 2017.

\bibitem{KLLLM2018}
Jon Kleinberg, Himabindu Lakkaraju, Jure Leskovec, Jens Ludwig, and Sendhil
  Mullainathan.
\newblock Human decisions and machine predictions.
\newblock {\em The Quarterly Journal of Economics}, 133(1), 2018.

\bibitem{Becker1957}
Gary Becker.
\newblock {\em The Economics of Discrimination}.
\newblock University of Chicago Press, 1957.

\bibitem{AltonjiBlank(99)}
Joseph Altonji and Rebecca Blank.
\newblock Race and gender in the labor market.
\newblock In Orley~C. Ashenfelter and David Card, editors, {\em Handbook of
  Labor Economics, Volume 3C}, pages 3143--3259. North Holland, 1999.

\bibitem{KnowlesPersicoTodd2001}
John Knowles, Nicola Persico, and Petra Todd.
\newblock Racial bias in motor vehicle searches: Theory and evidence.
\newblock {\em The Journal of Political Economy}, 109(1), 2001.

\bibitem{AnwarFang2006}
Shamena Anwar and Hanming Fang.
\newblock An alternative test of racial prejudice in motor vehicle searches:
  Theory and evidence.
\newblock {\em American Economic Review}, 96(1), 2006.

\bibitem{ArnoldDobbieYang2017}
David Arnold, Will Dobbie, and Crystal Yang.
\newblock Racial bias in bail decisions.
\newblock {\em The Quarterly Journal of Economics}, 133(4):1885–--1932, 2018.

\bibitem{CorbettDaviesGoel(18)}
Sam Corbett-Davies and Sharad Goel.
\newblock The measure and mismeasure of fairness: A critical review of fair
  machine learning.
\newblock Technical report, Stanford University Working Paper, 2018.

\bibitem{KallusZhou(18)}
Nathan Kallus and Angela Zhou.
\newblock Residual unfairness in fair machine learning from prejudiced data.
\newblock In {\em Proceedings of the 35th International Conference on Machine
  Learning}. 2018.

\bibitem{DeArteaga(18)}
Maria De-Arteaga, Artur Dubrawski, and Alexandra Chouldechova.
\newblock Learning under selective labels in the presence of expert
  consistency.
\newblock Technical report, arXiv preprint arXiv:1807.00905, 2018.

\bibitem{Cowgill2019}
Bo~Cowgill.
\newblock Bias and productivity in humans and machines.
\newblock Technical report, Columbia Business School Working Paper, 2019.

\bibitem{MadrasEtAl2019}
David Madras, Elliot Creager, Toniann Pitassi, and Richard Zemel.
\newblock Fairness through causal awareness: Learning causal latent-variable
  models for biased data.
\newblock In {\em Proceedings of the Conference on Fairness, Accountability,
  and Transparency}, FAT* '19, pages 349--358, 2019.

\bibitem{KannanEtAl(18)}
Sampath Kannan, Aaron Roth, and Juba Ziani.
\newblock Downstream effects of affirmative action.
\newblock Technical report, arXiv preprint arXiv 1808.09004, 2018.

\bibitem{Chander(17)}
Anupam Chander.
\newblock The racist algorithm.
\newblock {\em Michigan Law Review}, (6):1023--1046, 2017.

\bibitem{ShleiferEtAl(16)}
Pedro Bordalo, Katherine Coffman, Nicola Gennaioli, and Andrei Shleifer.
\newblock Stereotypes.
\newblock {\em The Quarterly Journal of Economics}, 131(4):1753--1794, 2016.

\bibitem{Li(17)}
Danielle Li.
\newblock Expertise vs. bias in evaluation: Evidence from the nih.
\newblock {\em American Economic Journal: Applied Economics}, 9(2), 2017.

\bibitem{KLR(16)}
Jon Kleinberg, Sendhil Mullainathan, and Manish Raghavan.
\newblock Inherent trade-offs in the fair determination of risk scores.
\newblock Technical report, arXiv preprint arXiv:1609.05807, 2016.

\bibitem{Cowgill2018}
Bo~Cowgill.
\newblock Bias and productivity in humans and machines: Theory and evidence.
\newblock Technical report, Columbia business School Working Paper, 2018.

\bibitem{RaghavanEtAl(19)}
Manish Raghavan, Solon Barocas, Jon Kleinberg, and Karen Levy.
\newblock Mitigating bias in algorithmic employment screening: Evaluating
  claims and practices.
\newblock Technical report, arXiv preprint arXiv: 1906.09208, 2019.

\bibitem{mullainathan2017does}
Sendhil Mullainathan and Ziad Obermeyer.
\newblock Does machine learning automate moral hazard and error?
\newblock {\em American Economic Review}, 107(5):476--80, 2017.

\bibitem{KLMS(18)}
Jon Kleinberg, Jens Ludwig, Sendhil Mullainathan, and Cass Sunstein.
\newblock Discrimination in the age of algorithms.
\newblock {\em Journal of Legal Analysis}, 80:1--62, 2018.

\bibitem{GillisSpiess(19)}
Talia Gillis and Jann Spiess.
\newblock Big data and discrimination.
\newblock {\em The University of Chicago Law Review}, 86:459--487, 2019.

\bibitem{KLMR(18)}
Jon Kleinberg, Jens Ludwig, Sendhil Mullainathan, and Ashesh Rambachan.
\newblock Algorithmic fairness.
\newblock {\em AEA Papers and Proceedings}, 108:22--27, 2018.

\bibitem{LiptonEtAl(18)}
Zachary Lipton, Alexandra Chouldechova, and Julian McAuley.
\newblock Does mitigating ml’s impact disparity require treatment disparity?
\newblock In S.~Bengio, H.~Wallach, H.~Larochelle, K.~Grauman, N.~Cesa-Bianchi,
  and R.~Garnett, editors, {\em Advances in Neural Information Processing
  Systems 31}, pages 8125--8135. 2018.

\bibitem{ChenEtAl2019}
Jiahao Chen, Nathan Kallus, Xiaojie Mao, Geoffry Svacha, and Madeleine Udell.
\newblock Fairness under unawareness: Assessing disparity when protected class
  is unobserved.
\newblock In {\em Proceedings of the Conference on Fairness, Accountability,
  and Transparency '19}, pages 339--348, 2019.

\bibitem{DattaEtAl(17)}
Anupam Datta, Matt Fredrikson, Gihyuk Ko, Piotr Mardziel, and Shayak Sen.
\newblock Proxy non-discrimination in data-driven systems.
\newblock Technical report, arXiv preprint arXiv:1707.08120, 2017.

\bibitem{GoelEtAl2016}
Sharad Goel, Justin Rao, and Ravi Shroff.
\newblock Precinct or prejudice? understanding racial disparities in new york
  city's stop-and-frisk policy.
\newblock {\em The Annals of Applied Statistics}, 10(1), 2016.

\bibitem{DworkEtAl(12)}
Cynthia Dwork, Moritz Hardt, Toniann Pitassi, Omer Reingold, and Richard Zemel.
\newblock Fairness through awareness.
\newblock {\em ITCS '12 Proceedings of the 3rd Innovations in Theoretical
  Computer Science Conference}, pages 214--226, 2012.

\end{thebibliography}

\newpage
\clearpage
\singlespacing
\appendix

\section{Proofs of Additional Results}

\subsection{Proof of Proposition \ref{prop:noise}}

The proof of Proposition \ref{prop:noise} uses the following lemma.

\begin{lem}
\label{lem:noise}
Suppose the police search individuals according to
    $$S^{noise}(X, U, R, \epsilon) = 1\Big(\mathbb{E}[Y | X, U, R] + \epsilon \geq c - \tau \cdot R\Big),$$
for a random prediction error $\epsilon$. Suppose that $\epsilon \indep U | X,R$ and the distribution of $\epsilon \,|\, X, R$ has an increasing hazard, i.e. $\frac{f(\epsilon | X, R)}{1 - F(\epsilon | X, R)} \mbox{ is increasing in } \epsilon$ for $f(\cdot \,|\, X,R)$ the conditional density function of $\epsilon$. Then, $\mu_{X,R,U} | \{S^{noise}=1, X, R=1\}$ has the monotone likelihood ratio property in $-\tau$, where $\mu_{X,R,U} = \mathbb{E}[Y | X, U, R]$ as before. 
\end{lem}

\noindent \textit{Proof}. The police choose $S^{noise} = 1$ if and only if $\mu_{X,R,U} + \epsilon \geq c - \tau \cdot R$, or equivalently, if and only if $\epsilon \geq c - \tau \cdot R - \mu_{X,R,U}$. Consider $\mu_1 < \mu_2$ in the support of $\mu_{X,R,U}$. Then, 
\begin{align*}
    &\dfrac{ \prob{\mu_{X,R,U} = \mu_1 | S^{noise} = 1, X , R} }{ \prob{\mu_{X,R,U} = \mu_2 | S^{noise} = 1, X , R} } \\
    &= \dfrac{ \prob{S^{noise}=1 | \mu_{X,R,U} = \mu_1 , X , R} }{ \prob{S^{noise}=1 | \mu_{X,R,U} = \mu_2 , X , R}  } \times \dfrac{ \prob{\mu_{X,R,U} = \mu_1 | X, R } / \prob{ S^{noise} =1 | X, R }}{\prob{\mu_{X,R,U} = \mu_2 | X, R } / \prob{ S^{noise} =1 | X, R }} \\
    &= \dfrac{ \prob{\epsilon \geq c - \tau \cdot R - \mu_1 | X , R} \cdot \prob{\mu_{X,R,U} = \mu_1 | X, R }  }{ \prob{\epsilon \geq c - \tau \cdot R - \mu_2 | X , R} \cdot \prob{\mu_{X,R,U} = \mu_2 | X, R }   } \\
    &= \dfrac{\left(1 -  F_{\epsilon | X,R}\left[ c - \tau \cdot R - \mu_1 \right] \right) \cdot \prob{\mu_{X,R,U} = \mu_1 | X, R }  }{\left(1 -  F_{\epsilon | X,R}\left[ c - \tau \cdot R - \mu_2 \right] \right)  \cdot \prob{\mu_{X,R,U} = \mu_2 | X, R }   }
\end{align*}
\noindent where the first equality follows from Bayes' Rule, the second equality uses the definition of $S^{noise}$ and the conditional independence of $\epsilon$ and $U$, and the third applies the definition of the CDF. Now, differentiating with respect to $-\tau$:

\begin{align*}
    & \dfrac{\partial}{\partial (-\tau)} \left( \dfrac{ \prob{\mu_{X,R,U} = \mu_1 | S^{noise} = 1, X , R} }{ \prob{\mu_{X,R,U} = \mu_2 | S^{noise} = 1, X , R} } \right) = \\
    & R \cdot \left( \dfrac{ f_{\epsilon | X,R}\left[ c - \tau \cdot R - \mu_1 \right] \left(1 - F_{\epsilon | X,R}\left[ c - \tau \cdot R - \mu_2 \right]  \right)   }{\left(1 -  F_{\epsilon | X,R}\left[ c - \tau \cdot R - \mu_2 \right] \right)^2} - \right. \\
    & \left.  \dfrac{ f_{\epsilon | X,R}\left[ c - \tau \cdot R - \mu_2 \right] \left(1 - F_{\epsilon | X,R}\left[ c - \tau \cdot R - \mu_1 \right]  \right)  }{\left(1 -  F_{\epsilon | X,R}\left[ c - \tau \cdot R - \mu_2 \right] \right)^2} \right) \times \dfrac{\prob{\mu_{X,R,U} = \mu_1 | X, R }}{\prob{\mu_{X,R,U} = \mu_2 | X, R }},
\end{align*}
\noindent Clearly, this derivative is zero if $R=0$. If $R=1$, the derivative is greater than or equal to zero if and only if 
\begin{align*}
    & f_{\epsilon | X,R}\left[ c - \tau \cdot R - \mu_1 \right] \left(1 - F_{\epsilon | X,R}\left[ c - \tau \cdot R - \mu_2 \right]  \right) \\
    &-  f_{\epsilon | X,R}\left[ c - \tau \cdot R - \mu_2 \right] \left(1 - F_{\epsilon | X,R}\left[ c - \tau \cdot R - \mu_1 \right]  \right) \geq 0
\end{align*}
\noindent or equivalently,
\begin{align}\label{eqn: comparison of hazards}
    \dfrac{f_{\epsilon | X,R}\left[ c - \tau \cdot R - \mu_1 \right]}{ 1 - F_{\epsilon | X,R}\left[ c - \tau \cdot R - \mu_1 \right] } \geq \dfrac{f_{\epsilon | X,R}\left[ c - \tau \cdot R - \mu_2 \right]}{ 1 - F_{\epsilon | X,R}\left[ c - \tau \cdot R - \mu_2 \right] }.
\end{align}
\noindent However, since $\mu_1 < \mu_2$, we have $c - \tau \cdot R - \mu_1 > c - \tau \cdot R - \mu_2$, and so (\ref{eqn: comparison of hazards}) holds if $\epsilon | X, R$ has increasing hazard. $\Box$

Returning to Proposition \ref{prop:noise}, we follow an argument that is analogous to the proof of Theorem \ref{theorem:comparative-static}. As before, define $\mu_{X,R,U} := \mathbb{E}[Y | X, R, U]$. Note that $S^{noise} = 1$ if and only if $\mu_{X, R, U} + \epsilon \geq c - \tau \cdot R$. Applying the law of iterated expectations, 
    \begin{align*}
        \mathbb{E}\left[ Y \,|\, X = x, R = r, S^{noise} = 1 \right] \overset{(1)}{=} \mathbb{E}\left[ \mathbb{E}\left[Y \,|\, X, R, U, \epsilon \right] \,|\, X = x, R = r, S^{noise} = 1 \right] \\
        \overset{(2)}{=} \mathbb{E}\left[ \mu_{X, R, U} \,|\, X = x, R = r, S^{noise} = 1 \right],
    \end{align*}
where (1) uses the law of iterated expectations and that $S^{noise}$ is simply a function of $X, U, R, \epsilon$ and (2) uses $\epsilon \indep Y \,| X, U, R$. The result then follows from Lemma \ref{lem:noise}.

\end{document}